\documentclass[journal]{IEEEtran}

\ifCLASSINFOpdf
\else
\fi

\usepackage{amsthm}
\usepackage{amsmath,amssymb}
\usepackage{amsfonts}
\usepackage{graphicx,subfigure}
\usepackage{float}
\usepackage{graphics}
\usepackage{booktabs}
\usepackage{multirow}
\usepackage{cite}
\usepackage{caption}
\usepackage[usenames, dvipsnames]{color}
\usepackage{graphicx} 
\usepackage{bm}
\newcommand{\EQ}{\begin{eqnarray}}
\newcommand{\EN}{\end{eqnarray}}
\newtheorem{assumption}{Assumption}

\usepackage{tabulary}
\newtheorem{thm}{Theorem}

\begin{document} 
%
\title{Fully Decoupled Neural Network Learning Using Delayed Gradients}

\author{
\IEEEauthorblockN{}
\IEEEauthorblockA{}
}

\author{Huiping Zhuang$^{a}$, Yi Wang$^{a}$, Qinglai Liu$^{a}$, Shuai Zhang$^{b}$, Zhiping Lin$^{a}$\\ 
	$^{a}$School of Electrical and Eletronic Engineering, Nanyang Technological University, Sinapore \\
	$^{b}$SenseTime, China \\
	\texttt{$^{a}$\{HUIPING001@e.,WANG1241@e.,liuql@,EZPLIN@\}ntu.edu.sg, $^{b}$zhangshuai@sensetime.com} 
}
\maketitle
\begin{abstract}
	Training neural networks with back-propagation (BP) requires a sequential passing of activations and gradients, which forces the network modules to work in a synchronous fashion. This has been recognized as the lockings (i.e., the forward, backward and update lockings) inherited from the BP. In this paper, we propose a fully decoupled training scheme using delayed gradients (FDG) to break all these lockings. The FDG splits a neural network into multiple modules and trains them independently and asynchronously using different workers (e.g., GPUs). We also introduce a gradient shrinking process to reduce the stale gradient effect caused by the delayed gradients. In addition, we prove that the proposed FDG algorithm guarantees a statistical convergence during training. Experiments are conducted by training deep convolutional neural networks to perform classification tasks on benchmark datasets, showing comparable or better results against the state-of-the-art methods as well as the BP in terms of both generalization and acceleration abilities. In particular, we show that the FDG is also able to train very wide networks (e.g., WRN-28-10) and extremely deep networks (e.g., ResNet-1202).
\end{abstract}

\section{Introduction}
In recent years, deep neural networks, e.g., convolutional neural network (CNN) \cite{lecun1998gradient} and recurrent neural network \cite{hochreiter1997long,cho2014learning}, have demonstrated great success in numerous highly complex tasks. Such success is built, to a great extent, on the ability to train extremely deep networks enabled by ResNet \cite{he2016deep} or other techniques with skip-connection-like structures \cite{zagoruyko2016wide,xie2017aggregated,huang2017densely,gastaldi2017shake}. Training networks with back-propagation (BP) \cite{werbos1974beyond} is a standard practice but it requires a complete forward and backward pass before the parameter update can be finished. This easily leads to inefficiency \cite{David2005Kickback} especially for training deeper networks, which is recognized as the lockings \cite{jaderberg2017decoupled} (i.e., forward, backward and update lockings) inherited from the standard BP. The existence of these lockings keeps the majority of the network on hold during the training, thereby compromising the efficiency.

In order to improve the efficiency, there have been a number of contributions on decoupling the training by splitting the network into multiple modules to facilitate model parallelization. With a common target for acceleration, the decoupled learning has several benefits over methods based on the data-parallel or mixed-parallel paradigm \cite{huang2018gpipe,jia2018beyond,sergeev2018horovod}. For instance, it avoids the performance loss for modules sensitive to batch size change, e.g., Batch Normalization \cite{ioffe2015batch}, and is able to parallelize the Recurrent neural networks (see \cite{jaderberg2017decoupled}). The decoupling techniques might be categorized into two groups: the backward-unlocking (BU) based methods and the local error learning (LEL) based methods.

The BU-based methods have access to the global information from the top layer and could break the backward locking. An additional benefit is that they often introduce no extra trainable parameters while enabling decoupling behaviors. Nonetheless, a full forward pass is still required before any parameter update. One important motivation for these techniques is to promote biological plausibility, which focuses on removing the weight symmetry and the gradient propagation from the BP. Feedback alignment (FA) \cite{lillicrap2016random} removes the weight symmetry by replacing symmetrical weights with random ones. Direct feedback alignment \cite{nokland2016direct} following the FA replaces the BP with a random projection and enables a simultaneous update for all layers. However, these biologically inspired approaches suffer from performance losses and are shown to scale poorly on more complex datasets \cite{bartunov2018assessing}. On the other hand, \textit{delayed gradients}  provide another solution of breaking the backward locking. The decoupled parallel BP using delayed gradients (DDG) \cite{huo2018decoupled} is able to train very deep (up to 110 layers) CNNs and shows no performance loss while reducing the training time. Since the DDG is still constrained by the forward locking, the acceleration is relatively limited even with multiple GPUs. The feature replay (FR) following the DDG also breaks the backward locking through recomputation, and it has been shown to perform even better than BP for several deep architectures with less memory consumption. However, the FR introduces more computational burden and thus is slower than the DDG.

The LEL-based methods use the local information and are more promising in terms of decoupling ability. This is because potentially they are able to fully decouple the neural network training. The full decoupling can be achieved by building auxiliary local loss functions to generate local error gradients, severing the gradient flow between the adjacent modules thereby training them asynchronously in parallel. The decoupled neural interface (DNI) proposed in \cite{jaderberg2017decoupled} is one of the pioneers exhibiting parallel training potential for neural networks. This technique utilizes a local neural network to generate synthetic gradients for the hidden layers so that the update could happen before completing either the forward or the backward pass. However, the DNI has been shown to learn poorly and even exhibit convergence problems in deeper networks \cite{huo2018decoupled}. In \cite{mostafa2018deep}, local classifiers with cross-entropy loss are adopted showing potentials to train the hidden layers simultaneously. It has been shown that the local classifier alone fails to match the performance of a standard BP. In \cite{nokland2019training}, a similarity measure combined with the local classifier is introduced to provide local error gradients. The mixed loss functions can produce classification performances comparable with or even better than the BP baselines but are currently tested only in VGG-like networks ($\le 13$ layers). Very recently, the depth problem of the LEL-based methods is alleviated by decoupled greedy learning (DGL) \cite{belilovsky2019decoupled}, which is able to train very deep networks ($\ge 100$ layers) while maintaining comparable performance against a standard BP. The common sacrifice that any LEL technique has to make is the introduction of extra trainable parameters imposed by the auxiliary networks. For instance, to match the standard BP, the local learning in \cite{nokland2019training} needs to train several times more parameters.

\begin{table} 
	\caption{Comparison with state-of-the-art methods in terms of lockings and auxiliary networks.}
	\label{table_unlock_aux}
	\resizebox{1\linewidth}{!}{%
		\begin{tabular}{cccccc}
			\toprule[0.3mm]
			Methods& DDG &FR& DNI   & DGL  &FDG (ours)\\ 
			\hline
			Lockings& Yes&Yes& No  & No & No\\
			Auxiliary networks & No & No& Yes  & Yes& No\\
			\hline
			\bottomrule[0.3mm]
		\end{tabular} 
	}
\end{table}

In summary, both BU-based and LEL-based methods can decouple the training of neural networks while showing potential in obtaining comparable performances against the standard BP. In comparison, the LEL-based methods lead in fully decoupling the network learning but introduce extra trainable parameters. The BU-based methods behave in the opposite way. In this paper, we propose a fully decoupled training scheme using delayed gradients (FDG) sharing both merits of the BU-based and the LEL-based techniques (see Table \ref{table_unlock_aux}). The main contributions of this work are as follows:

$\bullet$ We propose the FDG, a novel training technique  that breaks the forward, backward and update lockings without introducing extra trainable parameters. We also develop a gradient shrinking (GS) process that can reduce the stale gradient effect caused by utilizing the delayed gradients.

$\bullet$ We show that, in the ideal case, the FDG achieves a linear speedup w.r.t. to the number of workers.

$\bullet$ Theoretical analysis is provided showing that the proposed technique guarantees a statistical convergence.

$\bullet$ We conduct experiments by training deep CNNs and show that the proposed FDG produces comparable or better results compared with other state-of-the-art methods as well as the standard BP on benchmark datasets in terms of both generalization ability and computation time reduction.

Although we adopt delayed gradients like the DDG \cite{huo2018decoupled} and other asynchronous stochastic gradient descent (ASGD) methods \cite{dean2012large,lian2015asynchronous,zheng2017asynchronous}, the FDG is different from these techniques. The DDG only breaks the backward locking, but the FDG is able to break all the lockings, leading to a more efficient training. Different from the ASGD-based methods, the FDG trains the network by splitting it into modules handled by different workers while the ASGD-based methods  let each worker handle the entire network.

\section{Background}
In this section, we provide some background knowledge for training a feedforward neural network. The forward, backward and update lockings \cite{jaderberg2017decoupled} are also revisited.

Assume we need to train an $L$-layer network. The $l^{\text{th}}$ ($l\le L$) layer produces an activation $\bm{z}_{l} = A_{l}(\bm{z}_{l-1}; \bm{\theta}_l)$ by taking $\bm{z}_{l-1}$ as its input, where $A_l$ is an activation function and  $\bm{\theta}_{l}\in \mathbb{R}^{n_l}$ is a column vector representing the weights in layer $l$. The sequential generation of the activations constructs the \textit{forward locking}  since $\bm{z}_{l}$ will not be available before all the dependent activations are obtained. Let $\bm{\theta} = [\bm{\theta}_{1}^{T}, \bm{\theta}_{2}^{T}, ..., \bm{\theta}_{L}^{T}]^{T} \in \mathbb{R}^{\Sigma_{i=1}^{L}n_i}$ denote the parameter vector for the whole network. Assume $f$ is a loss function that maps a high-dimensional vector to a scalar. The learning of the feedforward network can then be summarized as the following optimization problem:
\begin{align}\label{eq_optimization}
\underset{\bm{\theta} = [\bm{\theta}_{1}^{T}, \bm{\theta}_{2}^{T}, ..., \bm{\theta}_{L}^{T}]^{T}}{\text{minimize}} \quad f_{\bm{x}}(\bm{\theta})
\end{align}
where $\bm{x}$ represents the input-label information (or training samples). We will drop the subscript $\bm{x}$ in \eqref{eq_optimization} in the rest of this paper for convenience: $f_{\bm{x}}(\bm{\theta})\rightarrow f(\bm{\theta})$.

The gradient descent algorithm is often used to solve \eqref{eq_optimization} by updating the parameter $\bm{\theta}$ iteratively. At step $t$, we have 
\begin{align}\label{eq_gd_batch}
\bm{\theta}^{t+1} = \bm{\theta}^{t} - \gamma_{t}\bm{\bar g}_{\theta}^{t}
\end{align}
or equivalently,
\begin{align}\label{eq_gd_update}
\bm{\theta}_{l}^{t+1} = \bm{\theta}_{l}^{t} - \gamma_{t}\bm{\bar g}_{\bm{\theta}_l}^{t}, \ l=1,...,L
\end{align}
where $\bm{\bar g}_{\theta}^{t}= [(\bm{\bar g}_{\bm{\theta}_1}^{t})^{T}, (\bm{\bar g}_{\bm{\theta}_2}^{t})^{T}, ..., (\bm{\bar g}_{\bm{\theta}_L}^{t})^{T}]^{T} \in \mathbb{R}^{\Sigma_{i=1}^{L}n_i}$ is the gradient vector obtained by
\begin{align}\label{eq_g_batch}
\bm{\bar g}_{\bm{\theta}_l}^{t} = \frac{\partial f(\bm{\theta}^{t})}{\partial\bm{\theta}_{l}^{t}}
\end{align}  
and $\gamma_{t}$ is the learning rate. If the training sample size is large, we apply stochastic gradient descent (SGD) as a replacement such that 
\begin{align}
\bm{g}_{\bm{\theta}_l}^{t} = \frac{\partial f_{\bm{x}_{t}}(\bm{\theta}^{t})}{\partial\bm{\theta}_{l}^{t}}
\end{align}
where $\bm{x}_{t}$ is a mini-batch of $\bm{x}$. Note the ``$\ \bar{} \ $''  has been removed to indicate the difference from \eqref{eq_g_batch}. Thus the parameter can be updated through
\begin{align}\label{eq_sgd_update}
\bm{\theta}_{l}^{t+1} = \bm{\theta}_{l}^{t} - \gamma_{t}\bm{g}_{\bm{\theta}_l}^{t}, \ l=1,...,L.
\end{align}  
Such a replacement is based on the realistic assumption of unbiased gradient as follows
\begin{align}\label{eq_expectation}
\mathrm{E}[\bm{g}_{\bm{\theta}_l}^{t}] = \bm{\bar g}_{\bm{\theta}_l}^{t}.
\end{align}
To obtain the gradient vectors, the BP (also known as the chain rule) can be employed. One could calculate the gradient in layer $l$ using the gradients back-propagated from layer $j$ and $i$ ($l<j<i$):
\begin{align}\label{eq_activation_chain1}
\bm{g}_{\bm{\theta}_l}^{t} =\frac{\partial f_{\bm{x}_{t}}(\bm{\theta}^{t})}{\partial\bm{\theta}_{l}^{t}} = \frac{\partial \bm{z}_{j}^{t}}{\partial \bm{\theta}_{l}^{t}}\frac{\partial f_{\bm{x}_{t}}(\bm{\theta}^{t})}{\partial\bm{z}_{j}^{t}} = \frac{\partial \bm{z}_{j}^{t}}{\partial \bm{\theta}_{l}^{t}}\bm{g}_{\bm{z}_j}^{t}
\end{align}
where 
\begin{align}\label{eq_activation_chain2}
\bm{g}_{\bm{z}_j}^{t} = \frac{\partial f_{\bm{x}_{t}}(\bm{\theta}^{t})}{\partial\bm{z}_{j}^{t}} = \frac{\partial \bm{z}_{i}^{t}}{\partial\bm{z}_{j}^{t}}\frac{\partial f_{\bm{x}_{t}}(\bm{\theta}^{t})}{\partial\bm{z}_{i}^{t}} = \frac{\partial \bm{z}_{i}^{t}}{\partial\bm{z}_{j}^{t}}\bm{g}_{\bm{z}_i}^{t}.
\end{align}
Note that we introduce $\bm{g}_{\bm{z}_j}^{t}$, i.e., the gradient vector w.r.t. activation $\bm{z}_j$, because it travels between modules as an important part of our proposed FDG. Equations \eqref{eq_activation_chain1} and \eqref{eq_activation_chain2} indicate a dependency of $\bm{g}_{\bm{\theta}_l}^{t}$ on $\bm{g}_{\bm{z}_j}^{t}$ and $\bm{g}_{\bm{z}_i}^{t}$. In other words, the gradient in layer $l$ remains unavailable until the gradient computations of all dependent layers are completed. This is also known as the \textit{backward locking}. In addition, the parameter update is not permitted before all modules complete executing the forward pass. This is recognized as the \textit{update locking}. In the following, we show that a full decoupling (i.e., the forward, backward and update unlockings) can be achieved.

\begin{figure*}
	\centering
	\includegraphics[width=1 \linewidth]{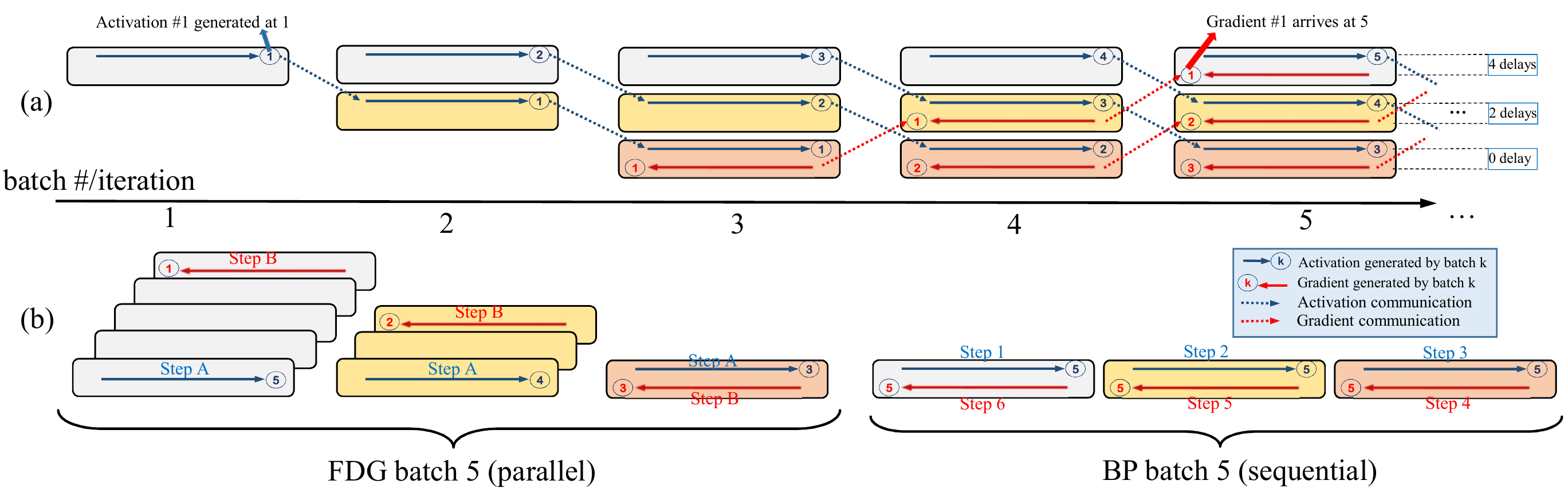}
	\caption{Illustration of the proposed FDG: a $K=3$ example. (a) By delaying the activations and gradients, the FDG allows the modules (in different colors) to be trained asynchronously in parallel. The communications among modules happen right before the iteration ends. We can easily see that there is a delay of $2(K-k)$ of gradients in module $k$ with a split of $K$. For instance, for $K=3$, module $k=1$ (gray) generates an activation at iteration 1 but the gradient of this batch arrives at iteration 5. (b) In the FDG, the backward pass (step B) is executed in the previously saved computation graph while the forward pass (step A) happens in the current one. This is how the FDG overcomes the sequential nature of the standard BP.}
	\label{fig:fdgalgorithm}
\end{figure*}

\section{Fully Decoupled Neural Network Learning}
In this section, we give the details of the proposed FDG. This technique provides a fully decoupled asynchronous learning algorithm with a gradient shrinking (GS) process which reduces the accuracy loss caused by the delayed gradients.

\subsection{The Proposed  FDG}
We first split the network into $K$ modules with each module containing a stack of layers. Accordingly, we split \{$1,\dots,L$\} into \{$q(1), q(2), \dots, q(K)$\} where $q(k) = \{m_k, m_{k}+1,...,m_{k+1}-1\}$ denotes the layer indices in module $k$.

As illustrated in Figure \ref{fig:fdgalgorithm}(a), during the decoupled learning, module $k$ is able to perform a forward and a backward pass using the delayed activation previously passed from module $k-1$ and the delayed gradient previously sent by module $k+1$. Note that all the modules undergo the same execution simultaneously, hence achieving the parallel training of different modules. After executing both passes, the gradient of the module input is passed to module $k-1$, while the module output is sent to module $k+1$ as its new input. This can be detailed by the following steps for module $k$ ($1<k<K$):

\noindent$\bullet$ \textbf{forward}: at iteration $t$, we feed the  input $\bm{z}_{m_{k}-1}^{t-k+1}$ (previously sent by module $k-1$) into module $k$ which produces a module output (activation) $\bm{z}_{m_{k+1}-1}^{t-k+1}$. We adopt $t-k+1$ instead of $t$ because at iteration $t$ we are using the delayed activation generated by batch $t-k+1$ (see Figure \ref{fig:fdgalgorithm}(a)).

\noindent$\bullet$ \textbf{backward}: at iteration $t$, we utilize the delayed gradient $\bm{g}_{\bm{z}_{m_{k+1}-1}}^{t-2K+k+1}$ previously received from module $k+1$ to resume the BP procedure. The superscript $t-2K+k+1$ is adopted because there is a delay of $2(K-k)$ of gradients (see Figure \ref{fig:fdgalgorithm}(a)) w.r.t. the forward pass  that uses batch $t-k+1$. Thus, for each layer ($m_{k} \le l \le m_{k+1}-1$), we obtain the gradient:
\begin{align}\label{eq_module_bp}
\bm{\hat g}_{\bm{\theta}_l}^{t-k+1} = \frac{\partial \bm{z}_{m_{k+1}-1}^{t-2K+k+1}}{\partial \bm{\theta}_{l}^{t-2K+k+1}}\bm{g}_{\bm{z}_{m_{k+1}-1}}^{t-2K+k+1}.
\end{align}
Note that \eqref{eq_module_bp} must utilize the delayed activation of batch ${t-2K+k+1}$. It is only reasonable to calculate the gradients based on the activations generated by the same training batch. Subsequently, the module can be updated through 
\begin{align}\label{eq_fdg_update}
\bm{\theta}_{l}^{t-k+2} = \bm{\theta}_{l}^{t-k+1} - \gamma_{t-k+1}\bm{\hat g}_{\bm{\theta}_l}^{t-k+1}.
\end{align}
After that, we save $\bm{g}_{\bm{z}_{m_k-1}}^{t-2K+k+1}$, the gradient of the module input, for communication.

\noindent$\bullet$ \textbf{communication}: at iteration $t$, send $\bm{g}_{\bm{z}_{m_k-1}}^{t-2K+k+1}$ to module $k-1$ and pass $\bm{z}_{m_{k+1}-1}^{t-k+1}$ to module $k+1$ as its new input. 

The aforementioned forward, backward and communication steps break all the lockings described in \cite{jaderberg2017decoupled}. Firstly, the global BP is broken into module-wise BP running in parallel, which achieves the \textit{backward unlocking}. Secondly, each of the split modules processes the training data from different batches, leading to an asynchronous module parallelization, hence the \textit{forward unlocking}. Finally, all the modules can be updated without waiting for other module to complete the forward pass, so the \textit{update unlocking} is also achieved.

Note that we utilize $t-k+1$ instead of $t$ in the update formula \eqref{eq_fdg_update} in correspondence to the activation index (e.g., $\bm{z}_{m_{k+1}-1}^{t-k+1}$) at iteration $t$. One could easily see that, for module $k$ at iteration $t+k-1$, \eqref{eq_fdg_update} can be equivalently shifted to
\begin{align}\label{eq_fdg_update2}
\bm{\theta}_{l}^{t+1} = \bm{\theta}_{l}^{t} - \gamma_{t}\bm{\hat g}_{\bm{\theta}_l}^{t}.
\end{align}
Adopting \eqref{eq_fdg_update2} over \eqref{eq_fdg_update} would benefit the subsequent convergence analysis in the next section. Let $d_{k,t} = t-2(K-k)$, we can further unpack $\bm{\hat g}_{\bm{\theta}_l}^{t}$ such that
\begin{align}
\resizebox{0.88\linewidth}{!}{$
	\bm{\hat g}_{\bm{\theta}_l}^{t} = \frac{\partial \bm{z}_{m_{k+1}-1}^{d_{k,t}}}{\partial \bm{\theta}_{l}^{d_{k,t}}}\frac{\partial f_{\bm{x}_{d_{k,t}}}(\bm{\theta}^{d_{k,t}})}{\partial \bm{z}_{m_{k+1}-1}^{d_{k,t}}} = \frac{\partial f_{\bm{x}_{d_{k,t}}}(\bm{\theta}^{d_{k,t}})}{\partial \bm{\theta}_{l}^{d_{k,t}}} = \bm{g}_{\bm{\theta}_l}^{d_{k,t}}
	$}
\end{align}
and rewrite \eqref{eq_fdg_update2} as
\begin{align}\label{eq_fdg_update3}
\bm{\theta}_{l}^{t+1} = \bm{\theta}_{l}^{t} - \gamma_{t}\bm{g}_{\bm{\theta}_l}^{d_{k,t}}
\end{align}
which is observed to take in delayed gradients with a delay of $2(K-k)$ compared with \eqref{eq_sgd_update}.

\noindent\textbf{Analysis of Speedup}: Assume a network is evenly split into $K$ modules. In the ideal case where other time consumptions such as communications are excluded, Table \ref{table_sp} shows that a linear ($K$-time) speedup can be achieved with the FDG, which is the highest among all the decoupling methods.

\begin{table}[h] 
	\caption{Ideal speedup of different decoupling methods for a network evenly split into $K$ modules. $\mathcal{T}_f$, $\mathcal{T}_b$ and $\mathcal{T}_{aux}$ denote the computation time executing the forward, the backward pass, and the auxiliary network, respectively.}
	\label{table_sp}
	\resizebox{1\linewidth}{!}{%
		\begin{tabular}{cccccc}
			\toprule[0.3mm]
			Methods& BP& DDG &FR& DNI  \& DGL  &FDG (ours)\\ 
			\midrule
			Time& $\mathcal{T}_f+\mathcal{T}_b$&$\mathcal{T}_f+\frac{\mathcal{T}_b}{K}$& $\mathcal{T}_f+\frac{\mathcal{T}_f+\mathcal{T}_b}{K}$  & $\frac{\mathcal{T}_f+\mathcal{T}_b}{K} + \mathcal{T}_{aux}$&$\frac{\mathcal{T}_f+\mathcal{T}_b}{K}$\\
			\bottomrule[0.3mm]
		\end{tabular} 
	}
\end{table}

\begin{figure}[b]
	\centering
	\includegraphics[width=1\linewidth]{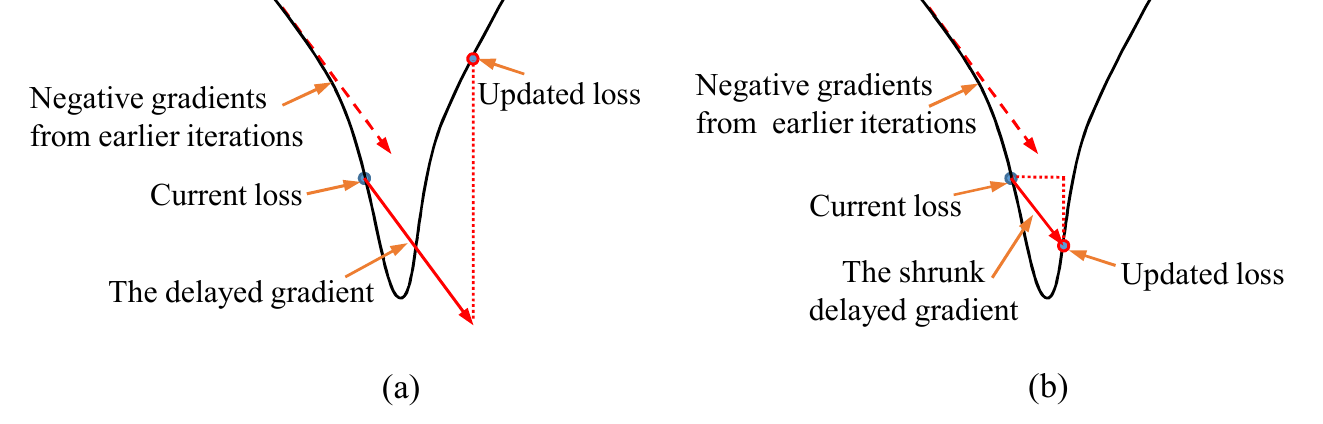}
	\caption{An intuitive interpretation for the GS process.}
	\label{fig:gs}
\end{figure}

\subsection{The Gradient Shrinking Process}
Using the delayed gradients enables model parallelization but could also lead to certain performance loss. This is a common phenomenon observed in algorithms with stale gradients \cite{chen2016revisiting}. To compensate the performance loss, we introduce a gradient shrinking (GS) process before back-propagating the delayed gradients through each module.

The GS process works in a straightforward manner. At iteration $t$, before executing BP in module $k$, we shrink the gradient by multiplying it with a shrinking factor $\beta$ ($0<\beta\le1$). This can be shown by modifying \eqref{eq_module_bp} as
\begin{align}\label{eq_gs_delay}
\bm{\hat g}_{\bm{\theta}_l}^{t-k+1} = \beta \frac{\partial \bm{z}_{m_{k+1}-1}^{t-2K+k+1}}{\partial \bm{\theta}_{l}^{t-2K+k+1}}\bm{ \hat{g}}_{\bm{z}_{m_{k+1}-1}}^{t-2K+k+1}.
\end{align}
The module is then updated through \eqref{eq_fdg_update}. If $\beta=1$, the GS process is not used. 

Note that in \eqref{eq_gs_delay}, unlike \eqref{eq_module_bp}, the received delayed gradient is denoted by $\bm{ \hat{g}}_{\bm{z}_{m_{k+1}-1}}^{t-2K+k+1}$ because the gradient in module $k$ is affected by multiple GS processes starting from the top module where the gradient is not shrunk. This can be illustrated by unpacking \eqref{eq_gs_delay}:
\begin{align}\nonumber
\bm{\hat g}_{\bm{\theta}_l}^{t-k+1} =& \beta^{2} \frac{\partial \bm{z}_{m_{k+1}-1}^{t-2K+k+1}}{\partial \bm{\theta}_{l}^{t-2K+k+1}}\frac{\partial \bm{z}_{m_{k+2}-1}^{t-2K+k+1}}{\partial \bm{z}_{m_{k+1}-1}^{t-2K+k+1}}\bm{\hat{g}}_{\bm{z}_{m_{k+2}-1}}^{t-2K+k+1}\\\nonumber
&\resizebox{!}{3mm}{$\mathbf{\vdots}$}\\\nonumber
=& \beta^{K-k} \frac{\partial \bm{z}_{m_{k+1}-1}^{t-2K+k+1}}{\partial \bm{\theta}_{l}^{t-2K+k+1}}\dots \frac{\partial \bm{z}_{m_{K}-1}^{t-2K+k+1}}{\partial \bm{z}_{m_{K-1}-1}^{t-2K+k+1}}\bm{g}_{\bm{z}_{m_{K}-1}}^{t-2K+k+1}\\\label{eq_gs_beta}
=& \beta^{K-k}\bm{g}_{\bm{\theta}_l}^{t-2K+k+1}
\end{align}
which shrinks the gradient by a factor of $\beta^{K-k}$ in module $k$. Equivalently, we have
\begin{align}\label{eq_gs_beta2}
\bm{\hat g}_{\bm{\theta}_l}^{t} = \beta^{K-k}\bm{g}_{\bm{\theta}_l}^{d_{k,t}}.
\end{align}

\begin{table}
	\centering
	\begin{tabulary}{\linewidth}{L}
		\toprule[0.3mm]
		\textbf{Algorithm I:} 	FDG (SGD)\\
		\toprule[0.3mm]
		\textbf{Required}: learning rate $\gamma_t$, number of split modules $K$, gradient shrinking factor $\beta$.\\
		$\bullet$ Split the network into $K$ modules and initialize them.\\
		\textcolor{blue}{\textbf{for}} $t=1,2,...,T$:\\
		\quad\textcolor{blue}{\textbf{Parallel for}} $k = 1,\dots,K$:\\
		\quad\quad$\bullet$ compute the shrunk delayed gradient in each layer\\
		\quad\quad $\bm{\hat g}_{\bm{\theta}_l}^{t-k+1} = \beta \frac{\partial \bm{z}_{m_{k+1}-1}^{t-2K+k+1}}{\partial \bm{\theta}_{l}^{t-2K+k+1}}\bm{\hat{g}}_{\bm{z}_{m_{k+1}-1}}^{t-2K+k+1}$ and gradient \\
		\quad\quad  of the  module input $\bm{\hat{g}}_{\bm{z}_{m_k-1}}^{t-2K+k+1}$.\\
		\quad\quad$\bullet$ update the module through \eqref{eq_fdg_update}.\\  
		\quad\quad$\bullet$ execute the forward pass to generate the module \\
		\quad\quad output $\bm{z}_{m_{k+1}-1}^{t-k+1}$.\\
		\quad\quad$\bullet$ pass $\bm{z}_{m_{k+1}-1}^{t-k+1}$ to module $k+1$ as its input.\\ 
		\quad\quad$\bullet$ send $\bm{\hat{g}}_{\bm{z}_{m_k-1}}^{t-2K+k+1}$ to module $k-1$.\\ 
		\quad {\color{blue}\bf End for}\\
		{\color{blue}\bf End for}\\
		\hline
		\bottomrule[0.3mm]
	\end{tabulary} 
\end{table}

The GS process works similarly by scaling the learning rate in the corresponding module, determining how much we should move towards the direction of the negative gradients. We can interpret this process in an intuitive way shown in Figure \ref{fig:gs}. The delayed gradients, especially with longer delays, would lead to deteriorated performance \cite{chen2016revisiting}.  Figure \ref{fig:gs}(a) shows a scenario where the delayed gradients cause the learning to miss the local minimum to a large margin. By using the shrunk delayed gradients, there is a better chance of reducing the stale gradient effect (see Figure \ref{fig:gs}(b) for an illustration). It is difficult to analyze how $\beta$ affects the network's ability to generalize. Instead, we will determine the value of $\beta$ empirically.

On the other hand, the FDG illustrated in Figure \ref{fig:fdgalgorithm}(b) also carries a message that the forward (step A) and the backward (step B) pass can be implemented separately. This means each module can execute the backward pass in advance. We discover that executing the backward pass first gives slightly better results (see Appendix A). Thus, for the experiments in this paper, the backward pass is implemented in higher priority. The proposed FDG with the GS process is summarized in Algorithm I with SGD optimizer.

\noindent\textbf{Comparison to DDG \cite{huo2018decoupled} and FR \cite{huo2018training}}: The DDG, the FR and the FDG all adopt delayed gradients. The DDG and FR address the backward locking only but the FDG breaks all the lockings. The fact that the DDG and FR require a full forward pass keeps the modules waiting before model parallelization, leading to possible poor resource utilizations (see GPU utilizations in the experiment section). Apart from the advanced unlocking properties, the introduced GS process is able to reduce the delayed gradient effect and even help the FDG surpass the standard BP (see the experiment section).

\section{Convergence Analysis}
In this section, we prove that the proposed FDG guarantees a statistical convergence. For convenience,  $\bm{\theta}^{t}, \bm{g}_{\bm{\theta}}^{t}$ and $\bm{\bar g}_{\theta}^{t}$ are rewritten in terms of modules as
\begin{align*}
\resizebox{0.03\linewidth}{!}{$\bm{\theta}^{t}$} =& \resizebox{0.37\linewidth}{!}{$[(\bm{\theta}_{q(1)}^{t})^{T},..., (\bm{\theta}_{q(K)}^{t})^{T}]^{T}$}, \resizebox{0.45\linewidth}{!}{$\bm{\theta}_{q(k)}^{t} =[(\bm{\theta}_{m_k}^{t})^{T},..., (\bm{\theta}_{m_{k+1}-1}^{t})^{T}]^{T}$}\\
\resizebox{0.03\linewidth}{!}{$\bm{g}_{\bm{\theta}}^{t}$}=&\resizebox{0.37\linewidth}{!}{$  [(\bm{g}_{\bm{\theta}_{q(1)}}^{t})^{T},..., (\bm{g}_{\bm{\theta}_{q(K)}}^{t})^{T}]^{T}$}, \resizebox{0.45\linewidth}{!}{$\bm{g}_{\bm{\theta}_{q(k)}}^{t} = [(\bm{g}_{\bm{\theta}_{m_k}}^{t})^{T},..., (\bm{g}_{\bm{\theta}_{m_{k+1}-1}}^{t})^{T}]^{T}$}\\
\resizebox{0.03\linewidth}{!}{$\bm{\bar g}_{\theta}^{t}$} =& \resizebox{0.37\linewidth}{!}{$[(\bm{\bar g}_{\bm{\theta}_{q(1)}}^{t})^{T},..., (\bm{\bar g}_{\bm{\theta}_{q(K)}}^{t})^{T}]^{T}$}, \resizebox{0.45\linewidth}{!}{$\bm{\bar g}_{\bm{\theta}_{q(k)}}^{t} = [(\bm{\bar g}_{\bm{\theta}_{m_k}}^{t})^{T},..., (\bm{\bar  g}_{\bm{\theta}_{m_{k+1}-1}}^{t})^{T}]^{T}$}.
\end{align*}

\begin{table*}
	\centering
	\caption{The Top 1 errors for various CNN structures on CIFAR-10 dataset under a split number $K=2$. Results with $^{*}$ are rerun using our training strategy. Overall, the rerun experiments give better results than those reported in the original papers.}
	\label{table_k_2_300}
	\resizebox{0.8\linewidth}{!}{%
		\begin{tabular}{lclcccc}
			\toprule[0.3mm]
			Architecture&\# params&\multicolumn{1}{c}{\text{BP}}&\text{DDG}  &\text{DGL} & \text{FR} &\text{FDG}\\ 
			\hline
			ResNet-20 &0.27M  &8.75\%/7.78\%$^{*}$&-&-&-&7.92\%($\beta$=1)/\textbf{7.23}\%($\beta$=0.2)\\ 
			ResNet-56 &0.46M&6.97\%/6.19\%$^{*}$ &6.89\%/6.63\%$^{*}$&6.77\%$^{*}$&6.07\%$^{*}$&6.20\%($\beta$=1)/\textbf{5.90}\%($\beta$=0.5)\\
			ResNet-110 &1.70M&6.43\%/5.79\%$^{*}$ &6.59\%/6.26\%$^{*}$&6.50\%/6.26\%$^{*}$&5.76\%$^{*}$&5.79\%($\beta$=1)/\textbf{5.73\%}($\beta$=0.2)\\
			ResNet-18  &11.2M&6.48\%/4.87\%$^{*}$&5.00\%$^{*}$&5.21\%$^{*}$&4.80\%$^{*}$&4.82\%($\beta$=1)/\textbf{4.79}\%($\beta$=0.8)\\
			ResNet-1202 &19.4M & 7.93\%/5.51\%$^{*}$  &-&-&-&5.50\%($\beta$=1)/\textbf{5.49}\%($\beta$=0.5)\\
			WRN-28-10 &36.5M&4.00\%/4.01\%$^{*}$ &4.05\%&4.12\%&3.87\%&4.13\%($\beta$=1)/\textbf{3.85}\%($\beta$=0.7)\\
			\hline
			\bottomrule[0.3mm]
		\end{tabular} 
	}
\end{table*}

\begin{assumption}\label{ass_continous}
	The gradients of the loss functions $f(\bm{\theta})$ and $f_{\bm{x}_{t}}(\bm{\theta})$ are  Lipschitz continuous. This means there exists a constant $L>0$ such that 
	\begin{align}
	||\bm{\bar g}_{\bm{\theta}}^{t_{1}} - \bm{\bar g}_{\bm{\theta}}^{t_{2}}||_2\le& L||\bm{\theta}^{t_{1}} - \bm{\theta}^{t_{2}}||_2\\
	||\bm{g}_{\bm{\theta}_{q(k)}}^{t_{1}} - \bm{g}_{\bm{\theta}_{q(k)}}^{t_{2}}||_2\le& L||\bm{\theta}_{q(k)}^{t_{1}} - \bm{\theta}_{q(k)}^{t_{2}}||_2.
	\end{align}
\end{assumption}
\begin{assumption}\label{ass_gradient_bound}
	The second moment of the stochastic gradient is bounded. This means $\forall t$, there exists a constant $M > 0$ such that:
	\begin{align}
	||\bm{g}_{\bm{\theta}}^{t}||_2^2\le M.
	\end{align}
\end{assumption}
Under Assumptions \ref{ass_continous} and \ref{ass_gradient_bound}, we can obtain the FDG's convergence property in the following theorem.

\begin{thm}\label{thm_convergence}
	Let Assumptions \ref{ass_continous} and \ref{ass_gradient_bound} hold. Assume that the learning rate is diminishing and $L\gamma_{t}\le1$. The proposed FDG in Algorithm I satisfies
	\begin{align}\label{eq_convergence}
	\mathrm{E}\Big[f(\bm{\theta}^{t+1})\Big] - \mathrm{E}\Big[f(\bm{\theta}^{t})\Big] \le -\frac{\gamma_{t}}{2}Z_{1} + \gamma_{t}^{2}Z_{2}
	\end{align}
	where
	\begin{align*}
	Z_{1} =& \sum\limits_{k=1}^{K}\beta^{K-k}||\bm{\bar g}_{\bm{\theta}_{q(k)}}^{t}||_2^2\\
	Z_{2} =& LM\frac{\beta^{K}-1}{\beta - 1} + LM\sum\limits_{k=1}^{K}\beta^{3(K-k)}(t-\mathrm{max}\{0,d_{k,t}\}).
	\end{align*}
\end{thm}

\begin{table*}[h]
	\centering
	\caption{The Top 1 errors for various CNN structures on CIFAR-100 dataset under $K=2$. Results with $^{*}$ are rerun using our training strategy. Overall, the rerun experiments give better results than those reported in the original papers.}
	\label{table_cifa100}
	\resizebox{0.8\linewidth}{!}{%
		\begin{tabular}{lcccccc}
			\toprule[0.3mm]
			Architecture&\# params&\multicolumn{1}{c}{\text{BP}}&\text{DDG}&DGL &\text{FR} &\text{FDG}\\ 
			\hline
			ResNet-56 &0.46M&30.21\%/27.68\%$^{*}$&29.83\%/28.44 \%$^{*}$&29.51\%$^{*}$&28.39\%$^{*}$& 27.87\%($\beta$=1)/\textbf{27.49}\%($\beta$=0.4)\\
			ResNet-110 &1.70M&28.10\%/25.82\%$^{*}$&28.61\%/27.16\%$^{*}$&26.80\%$^{*}$&26.31\%$^{*}$&25.73\%($\beta$=1)/\textbf{25.43}\%($\beta$=0.5)\\
			ResNet-18  &11.2M&22.35\%$^{*}$&22.74\%$^{*}$&22.24\%$^{*}$&22.88\%$^{*}$&22.78\%($\beta$=1)/\textbf{22.18}\%($\beta$=0.5)\\
			WRN-28-10 &36.5M&19.2\%/19.6\%$^{*}$ &-&-&-&20.28\%($\beta$=1)/\textbf{19.08}\%($\beta$=0.6)\\
			\hline
			\bottomrule[0.3mm]
		\end{tabular} 
	}
\end{table*}

As shown in Theorem \ref{thm_convergence}, the behavior of the expected loss value $\mathrm{E}[f(\bm{\theta}^{t+1})]$ is controlled by the learning rate $\gamma_{t}$. If the right side of \eqref{eq_convergence} is equal to or less than zero, i.e.,  
\begin{align*}
-\frac{\gamma_{t}}{2}Z_{1} + \gamma_{t}^{2}Z_{2}\le 0 \ => \ \gamma_{t} \le \text{min}\left\{\frac{1}{L}, \frac{Z_{1}}{2Z_{2}}\right\},
\end{align*}
the FDG guarantees the convergence statistically. The proof of Theorem \ref{thm_convergence} is provided in Appendix B.
\section{Experiments}\label{section_experiments}
In this section, we conduct experiments on CIFAR-10, CIFAR-100 \cite{krizhevsky2009learning} (benchmarked by the DDG and the FR) and Tiny-ImageNet datasets to compare the generalization and acceleration abilities among different decoupling methods. These experiments show that the proposed FDG provides comparable or better results than the standard BP as well as the state-of-the-art methods in terms of both generalization and acceleration performances.

\subsection{Comparison of Classification Performance}
\noindent\textbf{Implementation Details}: The experiments are conducted in the Pytorch platform with datasets pre-processed using standard data augmentation (i.e., random cropping, random horizontal flip and normalizing \cite{he2016deep,huang2017densely}). We use SGD optimizer with an initial learning rate of 0.1. The momentum and weight decay are set as 0.9 and $5\times 10^{-4}$ respectively. All the models are trained using a batch size of 128 for 300 epochs. The learning rate is divided by 10 at 150, 225 and 275 epochs. The test errors of all the experiments are reported at the \textit{last epoch} by the median of 3 runs. No validation set is used. For ResNet-110, ResNet-1202 and networks trained with $K=3,4$, we use $\gamma_{t} = 0.01$ to warm up the training for 3 epochs to avoid divergence.

We compare performances of five different methods, including the BP, the DDG \cite{huo2018decoupled}, the FR \cite{huo2018training}, the DGL \cite{belilovsky2019decoupled} and our proposed FDG. The DNI \cite{jaderberg2017decoupled} is not included as its performance has been shown to deteriorate severely with deeper networks \cite{huo2018decoupled}. The MLP-SR-aux in \cite{belilovsky2019decoupled} is adopted as the auxiliary network for the DGL.

\noindent\textbf{CIFAR-10}: We begin by reporting the classification results on the CIFAR-10 dataset, which is of 32x32 color images and includes 50000 training and 10000 testing samples with 10 classes. In this experiment, we split the original network at the center into two modules ($K=2$) and train them asynchronously and independently in 2 GPUs. To ensure fair comparisons among different methods, we \textit{rerun} the training using the BP, the DDG, the FR and the DGL with our training strategy.

The corresponding classification results are reported in Table \ref{table_k_2_300}. For ResNet-56, ResNet-110 and ResNet-18, networks trained by the FR give better generalization abilities over those trained by the DDG as claimed in \cite{huo2018training}. The FR also outperforms the DGL and even provides results slightly surpassing the BP baselines. The proposed FDG is validated by reporting the individual results with and without the GS process. Without the GS process, the FDG overtakes the DDG and the DGL, and achieves comparable results with the BP baselines. However, with a GS process, networks trained by the FDG are able to generalize better than their BP counterparts as well as those trained by the FR. 

Additionally, to show that the FDG is able to handle networks with various widths and depths, we provide the decoupled training for the shallower network (ResNet-20), the wider network (WRN-28-10) and the extremely deep network (ResNet-1202). All of these trained networks also generalize comparably to or better than those trained by the BP.

\noindent\textbf{CIFAR-100}: We now study the classification performance ($K=2$) on CIFAR-100, which contains the same number of training and testing samples as CIFAR-10 but with 100 classes. We again \textit{rerun} the experiments using BP and other methods with our training strategy. The performances are reported by the Top 1 error rates in Table \ref{table_cifa100}.   For ResNet-56, ResNet-110 and ResNet-18, we observe that, although overall the FR still overtakes the DDG and the DGL, it falls behind the BP. However, with the GS process, the FDG again beats the BP and other state-of-the-art methods. For our FDG, we provide the results for WRN-28-10, which also outperform the BP baseline with the help of the GS process.

\noindent\textbf{Tiny-ImageNet}: We finally report the performances of ResNet-18 and MobileNet v2 on Tiny-ImageNet, which is of 64x64 color images and has 200 classes, 100000 images for training 10000 images for testing. Similar to the CIFAR cases, the FDG also obtains comparable results to BP's. The Tiny-ImageNet experiments show that the proposed FDG can handle various input sizes.

\begin{table}
	\centering
	\caption{The Top 1 errors for ResNet-18 and MobileNet v2 on Tiny-ImageNet dataset under $K=2$.}
	\label{table_tiny}
	\resizebox{1\linewidth}{!}{%
		\begin{tabular}{cccc}
			\toprule[0.3mm]
			Architecture&\# params&\text{BP}&\text{FDG}\\ 
			\hline
			ResNet-18 & 11.2M & 38.32\%$^{*}$&38.58\%($\beta$=1)/\textbf{38.22}\%($\beta$=0.5)\\
			MobileNet v2 & 3.40M & \textbf{46.35\%}$^{*}$&46.36\%($\beta$=1)/46.44\%($\beta$=0.3)\\
			\hline
			\bottomrule[0.3mm]
		\end{tabular} 
	}
\end{table}

\begin{figure*}
	\centering
	\includegraphics[width=1\linewidth]{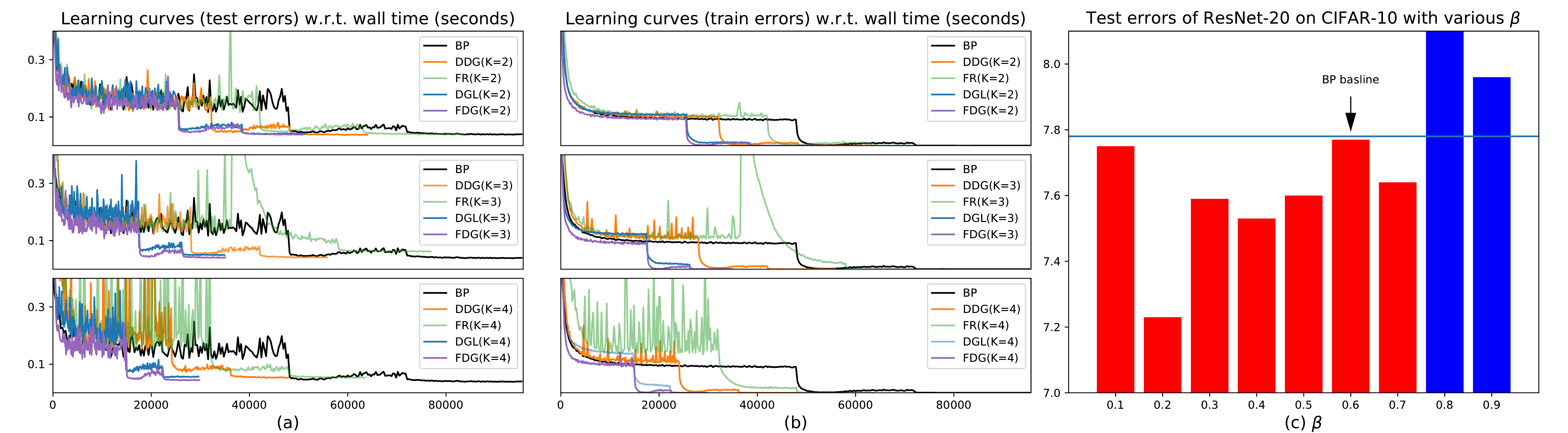}
	\caption{(a)-(b) Learning curves of the BP, the DDG, the FR, the DGL and the FDG for WRN-28-10 with $K=2,3,4$. (c) The error rates of ResNet-20 trained by the FDG with various $\beta$ values.}
	\label{fig:fdgsplittwo}
\end{figure*}

\noindent\textbf{The Impact of the GS Process:} The GS process with a proper $\beta$ could enhance the FDG's generalization ability. We now empirically evaluate the impact of the GS process by experimenting with various values of the shrinking factor $\beta$. This evaluation is conducted by training the ResNet-20 on CIFAR-10 dataset. The bar chart in Figure \ref{fig:fdgsplittwo}(c) reports the Top 1 error rates. We notice that the results for the proposed FDG are able to surpass the BP baseline with a small effort of tunning the $\beta$. This also shows that the GS process does enhance a network's ability to generalize.

\noindent\textbf{More Split Modules:} In this experiment, we study the performance of ResNet-56 and WRN-28-10 on CIFAR-10 by splitting them into $K=3$ and $K=4$ modules with each module trained in an independent GPU. The results are shown in Table \ref{table_many_modules} where we list the test errors with $K=2, 3, 4$. It becomes noticeable that more split modules have caused all these methods to lose accuracy. However, we also observe that the GS process allows the classification performances to be restored to the level of the BP baseline. The improved performances indicate that the GS process plays an essential role in reducing the stale gradient effect, which becomes even more significant as $K$ increases. 

\begin{table}[h]
	\centering
	\caption{The Top 1 errors for ResNet-56 on CIFAR-10 dataset under a split number $K=2, 3, 4$.}
	\label{table_many_modules}
	\resizebox{1\linewidth}{!}{%
		\begin{tabular}{cccccc}
			\toprule[0.3mm]
			&\multicolumn{1}{c}{\text{BP}}&\text{DDG}&DGL &\text{FR} &\text{FDG}\\ 
			\hline
			ResNet-56 ($K=2$)&6.19\%&6.60\%$^{*}$&6.77\%$^{*}$&6.07\%$^{*}$&6.20\%($\beta$=1)/\textbf{5.90}\%($\beta$=0.5)\\
			ResNet-56 ($K=3$)&6.19\%&6.50\%$^{*}$&8.88\%$^{*}$&6.33\%$^{*}$&6.40\%($\beta$=1)/\textbf{6.08}\%($\beta$=0.2)\\
			ResNet-56 ($K=4$)&6.19\%&6.61\%$^{*}$&9.65\%$^{*}$& 6.48\%$^{*}$ &6.83\%($\beta$=1)/\textbf{6.14}\%($\beta$=0.3)\\
			\hline
			WRN-28-10 ($K=2$)&4.01\%&4.05\%$^{*}$&4.12\%$^{*}$&3.87\%$^{*}$&4.13\%($\beta$=1)/\textbf{3.85}\%($\beta$=0.7)\\
			WRN-28-10 ($K=3$)&\textbf{4.01}\%&4.12\%$^{*}$&4.91\%$^{*}$&6.16\%$^{*}$&4.19\%($\beta$=1)/4.07\%($\beta$=0.5)\\
			WRN-28-10 ($K=4$)&\textbf{4.01}\%&6.61\%$^{*}$&5.64\%$^{*}$& 5.39\%$^{*}$ &6.50\%($\beta$=1)/4.42\%($\beta$=0.5)\\
			\hline
			\bottomrule[0.3mm]
	\end{tabular} 	}
\end{table}

As an example, the learning curves of the WRN28-10 with various $K$ are also plotted in Fig. \ref{fig:fdgsplittwo}(a)-(b), where we can see that the fully decoupled methods (DGL and FDG) are much faster than the BU-based methods. The speedup comparisons are detailed in the following subsection.

\subsection{Comparison of Acceleration Performance}
We conduct experiments WRN-28-10 on CIFAR-10 and ResNet-101 on ImageNet with $K=2,3,4$ to compare the acceleration abilities among the decoupling techniques. To ensure fair comparisons, we reimplement the DDG, the FR and the DGL in our framework adopting the identical communication protocols. In particular, we adopt a pipeline parallelization for DGL, a pipeline version of the sequential DGL \cite{belilovsky2019decoupled} with each worker handling one specific module. This is to encourage a fair comparison of speedup among all the methods by allowing each worker to handle one module alone. The comparisons are done by reporting the number of images processed per second averaged in 10 seconds when the training is stabilized. The training speed as well as the GPU utilization are reported in Table \ref{table_acceleration}. These experiments are conducted using a server with Intel Xeon E5-1680v4 CPU and RTX 1080Ti GPUs. Note that the experiments are evaluated \textit{without high-speed interconnect among GPUs}, so the speedup potentials could not be fully revealed under our current hardware settings.

\begin{table*}
	\caption{Comparisons of speed (images/s) among different training methods on CIFAR-10 and ImageNet with GPU utilizations inside the brackets. For instance, 601.60 (71\%,94\%,80\%,61\%), 3.20$\times$ means 601.60 images per second with 4 GPUs whose utilizations are 71\%, 94\%, 80\% and 61\% respectively and a 3.2$\times$ speedup over BP. The DGL performs better for acceleration due to less communication cost but with obvious accuracy loss.}
	\label{table_acceleration}
	\resizebox{1\linewidth}{!}{%
		\begin{tabular}{ccc|cc|cc}
			\toprule[0.3mm]
			\hline
			&\multicolumn{2}{c|}{$K=2$}&\multicolumn{2}{c|}{$K=3$}&\multicolumn{2}{c}{$K=4$}\\
			\hline
			&WRN-28-10 (CIFAR-10) & ResNet-101 (ImageNet)&WRN-28-10 (CIFAR-10) & ResNet-101 (ImageNet)&WRN-28-10 (CIFAR-10) & ResNet-101 (ImageNet)\\ 
			\hline
			BP & 188.16 (99\%)&104.3 (98\%)&188.16 (99\%)&104.3 (98\%)&188.16 (99\%)&104.3 (98\%)\\
			DDG&280.32 (81\%,69\%), 1.49$\times$&155.5 (75\%,84\%), 1.49$\times$&312.28 (57\%,64\%,51\%), 1.66$\times$&176.8 (61\%,58\%,45\%), 1.70$\times$  &373.76 (90\%,60\%,69\%,35\%), 1.99$\times$ &189.0 (49\%,52\%,40\%,49\%), 1.81$\times$\\
			FR &215.04 (80\%,60\%), 1.14$\times$&125.0 (70\%,67\%), 1.20$\times$& 232.96 (51\%,68\%,70\%), 1.24$\times$&136.2 (76\%,70\%,39\%), 1.31$\times$  &281.60 (59\%,62\%,35\%,31\%), 1.50$\times$ &152.3 (51\%,61\%,37\%,48\%), 1.46$\times$\\
			DGL &349.44 (99\%,85\%), 1.86$\times$ &\textbf{194.6 (97\%,87\%), 1.89$\times$}& \textbf{514.56 (97\%,94\%,83\%), 2.73$\times$} &\textbf{273.0 (96\%,92\%78\%), 2.62$\times$}  &\textbf{611.84 (82\%,95\%,82\%,67\%), 3.25$\times$}&\textbf{332.5 (93\%,87\%,87\%,71\%), 3.19$\times$}\\
			FDG &\textbf{354.56 (98\%,87\%), 1.88$\times$} &175.0 (84\%,80\%), 1.88$\times$ &512.00 (85\%,93\%,75\%), 2.72$\times$&217.0 (74\%,63\%,62\%), 2.10$\times$  &601.60 (71\%,94\%,80\%,61\%), 3.20$\times$ &234.3 (60\%,68\%,59\%,54\%), 2.30$\times$\\
			\hline
			\bottomrule[0.3mm]
		\end{tabular} 
		
	}
	\textbf{Note}: the experiments are evaluated without high-speed interconnect among GPUs.
\end{table*}

\noindent\textbf{Comparisons among Decoupling Methods}: In Table \ref{table_acceleration}, for the WRN-28-10 (CIFAR-10) case, in general the proposed FDG obtains impressive acceleration results compared with other decoupling methods. For $K=2$, our FDG slightly outperforms the DGL and achieves a \textbf{1.88}$\times$ speedup compared with the BP! For $K=3$, an impressive \textbf{2.72}$\times$ speedup is also achieved! These accelerations, which are much faster than the DDG and FR, are very close to the linear speedup indicated in Table \ref{table_sp}, and are comparable to a decent data parallelization benchmark \cite{sergeev2018horovod} with optimized communication protocols. This is not surprising since the proposed FDG does not introduce any extra computation. For $K=4$, a \textbf{3.20}$\times$ speedup is obtained with the averaged GPU utilizations below 90\%. Note that the FDG is implemented with vanilla communication protocols, so the speedup is expected to be further improved with better hardware setting or more efficient communication protocols. Since the DGL does not need to transfer the gradients back to other modules, it becomes slightly faster due to less communication cost, but it comes with a loss of accuracy. Our vanilla communication implementation for FDG will become costly if a larger amount of data is transferred. This can be shown in the following ImageNet speedup experiments.

For the ResNet-101 (ImageNet) case, the FDG achieves a \textbf{1.68}$\times$ speedup ($K=2$), which is visibly slower than that in the WRN-28-10 experiment. This slow-down is normal due to the lack of high-speed connections among GPUs (e.g., a drop from 1.8$\times$ to 1$\times$ without high-speed interconnect in \cite{huang2018gpipe}). The slow-down becomes more significant for $K=3,4$ where only \textbf{2.1}$\times$ and \textbf{2.3}$\times$ speedups (still faster than the DDG and FR) are achieved. The sole difference from the WRN-28-10 case is that the ImageNet case needs to transfer much larger tensors across GPUs. Without high-speed communication bridges among GPUs (e.g., NVLink \cite{foley2017ultra}), the communication cost becomes more dominant with larger data for transmission and more GPUs involved.

On the other hand, we also find that training the networks using fully decoupled methods (i.e., FDG and DGL) usually give much higher GPU utilizations than those using the BU-based methods (i.e., DDG and FR). This is because the modules trained by the fully decoupled methods are always active with much less waiting overhead, unlike the BU-based methods which require a forward pass before parallelization.

\noindent\textbf{Suggestion for Further Acceleration}: As shown in Table \ref{table_acceleration}, the proposed FDG encounters slow-down in training larger datasets. There are several methods that can be adopted to improve the acceleration. Firstly, a high-speed physical GPU bridge (e.g., NVLink \cite{foley2017ultra}) should boost the acceleration significantly by taking away the expensive communication cost. Note this is not an unrealistic request as many distributed methods \cite{huang2018gpipe,sergeev2018horovod} consider the high-speed interconnect a default setting to fully show the speedup performance. Secondly, the module splitting strategy could be improved to ensure an equal computation load for each worker. Finally, more efficient communication protocols can be developed to reduce the communication burden. We will consider these attempts in future work.

\section{Conclusion}
In this paper, we propose a fully decoupled method using the delayed gradients (FDG) to break the forward, backward and update lockings for neural network learning. The breaking of these lockings leads to a module-wise parallelization, which enables the FDG to achieve up to a linear speedup in the ideal case. To enhance the FDG, we introduce the gradient shrinking process that has been shown to improve a network's ability to generalize. Theoretical analysis shows that the proposed FDG guarantees a statistical convergence. Our experiments on the CNNs indicate that the FDG outperforms the state-of-the-art methods and even overtakes the standard BP while achieving a significant acceleration (e.g., 1.88x with 2 GPUs and 2.72x with 3 GPUs). Our method also succeeds in training very wide networks as well as extremely deep networks. The experiments reveal that the FDG could potentially benefit from a more efficient communication process, specially for large-scale datasets.

{\small
	\bibliographystyle{ieee_fullname}
	\bibliography{DG}
}

\newpage
\onecolumn
\section*{ Appendix A: Some Primary Experiments}
We discover that executing the backward pass first (FDG-backward) could obtain slightly better results than that executing the forward pass first (FDG-forward). This can be shown in Table \ref{table_appendix1} by conducting some primary experiments for training ResNet-20 and ResNet-56 on CIFAR-10. The training strategy can be found in the experiment section.
\begin{table*}[h]
	\centering
	\caption{Some primary results for training ResNet-20 and ResNet-56 on CIFAR-10 dataset under a split number $K$=2.}
	\label{table_appendix1}
	\begin{tabular}{lccc}
		\toprule[0.3mm]
		Architecture&\# params&\text{FDG-forward} &\text{FDG-backward}\\ 
		\hline
		ResNet-20 &0.27M  &8.03\%($\beta$=1)/7.57\%($\beta$=0.2)&7.92\%($\beta$=1)/\textbf{7.23}\%($\beta$=0.2)\\ 
		ResNet-56 &0.46M  &6.20\%($\beta$=1)/5.94\%($\beta$=0.5)&6.20\%($\beta$=1)/\textbf{5.90}\%($\beta$=0.5)\\
		\hline
		\bottomrule[0.3mm]
	\end{tabular} 
\end{table*}

\section*{ Appendix B: Proof to Theorem 1}
\begin{proof}
	According to Assumption 1, the following inequality holds:
	\begin{align}\label{eq_continous}
	f(\bm{\theta}^{t+1}) \le f(\bm{\theta}^{t}) + (\bm{\bar g}_{\theta}^{t})^{T}(\bm{\theta}^{t+1} - \bm{\theta}^{t}) + \frac{L}{2}\Big\lVert\bm{\theta}^{t+1} - \bm{\theta}^{t}\Big\lVert_2^2.
	\end{align}
	According to (12) and (17), \eqref{eq_continous} can be rewritten as
	\begin{align}\nonumber
	f(\bm{\theta}^{t+1}) &\le f(\bm{\theta}^{t}) - \gamma_{t}\sum\limits_{k=1}^{K}\beta^{K-k}(\bm{\bar g}_{\bm{\theta}_{q(k)}}^{t})^{T}\bm{g}_{\bm{\theta}_{q(k)}}^{d_{k,t}} + \frac{L\gamma_{t}^{2}}{2}\sum\limits_{k=1}^{K}\Big\lVert\beta^{K-k}\bm{g}_{\bm{\theta}_{q(k)}}^{d_{k,t}}\Big\lVert_2^{2}\\\label{eq_appendix2}
	&\le f(\bm{\theta}^{t}) - \gamma_{t}\sum\limits_{k=1}^{K}\beta^{K-k}(\bm{\bar g}_{\bm{\theta}_{q(k)}}^{t})^{T}\bm{g}_{\bm{\theta}_{q(k)}}^{d_{k,t}}+ \frac{L\gamma_{t}^{2}}{2}\sum\limits_{k=1}^{K}\beta^{K-k}\Big\lVert\bm{g}_{\bm{\theta}_{q(k)}}^{d_{k,t}}\Big\lVert_2^{2}
	\end{align}
	where the second inequality is due to $0<\beta\le 1$. \eqref{eq_appendix2} establishes a module-wise relationship between the loss functions at $t+1$ and $t$. To prove the statistical convergence, our goal is to show that the expectation of the summation of the second and the third term is bounded. To this end, \eqref{eq_appendix2} can be further developed such that
	\begin{align}\nonumber
	f(\bm{\theta}^{t+1}) &\le f(\bm{\theta}^{t})- \gamma_{t}\sum\limits_{k=1}^{K}\beta^{K-k}(\bm{\bar g}_{\bm{\theta}_{q(k)}}^{t})^{T}\bm{g}_{\bm{\theta}_{q(k)}}^{d_{k,t}} + \frac{L\gamma_{t}^{2}}{2}\sum\limits_{k=1}^{K}\beta^{K-k}\Big\lVert\bm{g}_{\bm{\theta}_{q(k)}}^{d_{k,t}}\Big\lVert_2^{2}\\\nonumber
	&= f(\bm{\theta}^{t})- \gamma_{t}\sum\limits_{k=1}^{K}\beta^{K-k}(\bm{\bar g}_{\bm{\theta}_{q(k)}}^{t})^{T}(\bm{g}_{\bm{\theta}_{q(k)}}^{d_{k,t}} - \bm{\bar g}_{\bm{\theta}_{q(k)}}^{t} + \bm{\bar g}_{\bm{\theta}_{q(k)}}^{t})+ \frac{L\gamma_{t}^{2}}{2}\sum\limits_{k=1}^{K}\beta^{K-k}\Big\lVert\bm{g}_{\bm{\theta}_{q(k)}}^{d_{k,t}} - \bm{\bar g}_{\bm{\theta}_{q(k)}}^{t} + \bm{\bar g}_{\bm{\theta}_{q(k)}}^{t}\Big\lVert_2^{2}\\\nonumber
	&=f(\bm{\theta}^{t}) - \gamma_{t}\sum\limits_{k=1}^{K}\beta^{K-k}\Big\lVert_2^{2} \bm{\bar g}_{\bm{\theta}_{q(k)}}^{t}\Big\lVert_2^{2} -\gamma_{t}\sum\limits_{k=1}^{K}\beta^{K-k}(\bm{\bar g}_{\bm{\theta}_{q(k)}}^{t})^{T}(\bm{g}_{\bm{\theta}_{q(k)}}^{d_{k,t}} - \bm{\bar g}_{\bm{\theta}_{q(k)}}^{t}) + \frac{L\gamma_{t}^{2}}{2}\sum\limits_{k=1}^{K}\beta^{K-k}\Big\lVert \bm{\bar g}_{\bm{\theta}_{q(k)}}^{t}\Big\lVert_2^{2}\\\nonumber
	& + \frac{L\gamma_{t}^{2}}{2}\sum\limits_{k=1}^{K}\beta^{K-k}\Big\lVert\bm{g}_{\bm{\theta}_{q(k)}}^{d_{k,t}} - \bm{\bar g}_{\bm{\theta}_{q(k)}}^{t} \Big\lVert_2^{2} + \frac{L\gamma_{t}^{2}}{2}\sum\limits_{k=1}^{K}\beta^{K-k}(\bm{\bar g}_{\bm{\theta}_{q(k)}}^{t})^{T}(\bm{g}_{\bm{\theta}_{q(k)}}^{d_{k,t}} - \bm{\bar g}_{\bm{\theta}_{q(k)}}^{t})\\\label{eq_appendix_3}
	&= f(\bm{\theta}^{t}) - (\gamma_{t} - \frac{L\gamma_{t}^{2}}{2})\sum\limits_{k=1}^{K}\beta^{K-k}\Big\lVert\bm{\bar g}_{\bm{\theta}_{q(k)}}^{t}\Big\lVert_2^2 + \tilde{Q}_1 + \tilde{Q}_2.
	\end{align}
	where
	\begin{align*}
	\tilde{Q}_1 &= (\frac{L\gamma_{t}^{2}}{2} - \gamma_{t})\sum\limits_{k=1}^{K}\beta^{K-k}(\bm{\bar g}_{\bm{\theta}_{q(k)}}^{t})^{T}(\bm{g}_{\bm{\theta}_{q(k)}}^{d_{k,t}} - \bm{\bar g}_{\bm{\theta}_{q(k)}}^{t}) \\
	\tilde{Q}_2 &=  \frac{L\gamma_{t}^{2}}{2}\sum\limits_{k=1}^{K}\beta^{K-k}\Big\lVert\bm{g}_{\bm{\theta}_{q(k)}}^{d_{k,t}} - \bm{\bar g}_{\bm{\theta}_{q(k)}}^{t} \Big\lVert_2^{2}.
	\end{align*}
	Subsequently, the expectation of $\tilde{Q}_1$ is bounded by
	
	\begin{align*}
	\mathrm{E}[\tilde{Q}_1] =& \frac{L\gamma_{t}^{2}}{2}\mathrm{E}\Big[\sum\limits_{k=1}^{K}\beta^{K-k}\Big\lVert\bm{g}_{\bm{\theta}_{q(k)}}^{d_{k,t}} - \bm{\bar g}_{\bm{\theta}_{q(k)}}^{t}\Big\lVert_{2}^{2}\Big]\\
	=& \frac{L\gamma_{t}^{2}}{2}\mathrm{E}\Big[\sum\limits_{k=1}^{K}\beta^{K-k}\Big\lVert\bm{g}_{\bm{\theta}_{q(k)}}^{d_{k,t}} -\bm{\bar g}_{\bm{\theta}_{q(k)}}^{d_{k,t}} - \bm{\bar g}_{\bm{\theta}_{q(k)}}^{t} + \bm{\bar g}_{\bm{\theta}_{q(k)}}^{d_{k,t}} \Big\lVert_{2}^{2}\Big]\\
	\le & L\gamma_{t}^{2}\mathrm{E}\Big[\sum\limits_{k=1}^{K}\beta^{K-k}\Big\lVert\bm{g}_{\bm{\theta}_{q(k)}}^{d_{k,t}} -\bm{\bar g}_{\bm{\theta}_{q(k)}}^{d_{k,t}}\Big\lVert_{2}^{2}\Big] +   L\gamma_{t}^{2}\sum\limits_{k=1}^{K}\beta^{K-k}\Big\lVert\bm{\bar g}_{\bm{\theta}_{q(k)}}^{d_{k,t}} - \bm{\bar g}_{\bm{\theta}_{q(k)}}^{t}\Big\lVert_{2}^{2}\\
	=& L\gamma_{t}^{2}\sum\limits_{k=1}^{K}\beta^{K-k}\mathrm{E}\Big[\Big\lVert\bm{g}_{\bm{\theta}_{q(k)}}^{d_{k,t}} -\bm{\bar g}_{\bm{\theta}_{q(k)}}^{d_{k,t}}\Big\lVert_{2}^{2}\Big] +   L\gamma_{t}^{2}\sum\limits_{k=1}^{K}\beta^{K-k}\Big\lVert\bm{\bar g}_{\bm{\theta}_{q(k)}}^{d_{k,t}} - \bm{\bar g}_{\bm{\theta}_{q(k)}}^{t}\Big\lVert_{2}^{2}\\
	\le & L\gamma_{t}^{2}\sum\limits_{k=1}^{K}\beta^{K-k}\mathrm{E}\Big[\Big\lVert\bm{g}_{\bm{\theta}_{q(k)}}^{d_{k,t}}\Big\lVert_{2}^{2}\Big] +   L\gamma_{t}^{2}\sum\limits_{k=1}^{K}\beta^{K-k}\Big\lVert\bm{\bar g}_{\bm{\theta}_{q(k)}}^{d_{k,t}} - \bm{\bar g}_{\bm{\theta}_{q(k)}}^{t}\Big\lVert_{2}^{2}\\
	\le & L\gamma_{t}^{2}M\sum\limits_{k=1}^{K}\beta^{K-k} + L\gamma_{t}^{2}\sum\limits_{k=1}^{K}\beta^{K-k}\Big\lVert\bm{\bar g}_{\bm{\theta}_{q(k)}}^{d_{k,t}} - \bm{\bar g}_{\bm{\theta}_{q(k)}}^{t}\Big\lVert_{2}^{2}\\
	=&  L\gamma_{t}^{2}M\frac{\beta^{K}-1}{\beta - 1} + L\gamma_{t}^{2}\tilde{P}_1
	\end{align*}
	where the first inequality follows from $\lVert\bm{x} + \bm{y}\lVert_{2}^{2}\le 2\lVert\bm{x}\lVert_2^2 + 2\lVert\bm{y}\lVert_2^2$, the second one is from the unbiased property in (7) such that $\mathrm{E}[\lVert\epsilon - \mathrm{E}[\epsilon]\lVert_{2}^{2}]\le\mathrm{E}[\lVert\epsilon\lVert_{2}^{2}] -  \lVert\mathrm{E}[\epsilon]\lVert_{2}^{2}\le \mathrm{E}[\lVert\epsilon\lVert_{2}^{2}]$, the third one follows from Assumption 2, and the $\tilde{P}_1$ can be bounded by 
	\begin{align*}
	\tilde{P}_1 =& \sum\limits_{k=1}^{K}\beta^{K-k}\Big\lVert\bm{\bar g}_{\bm{\theta}_{q(k)}}^{d_{k,t}} - \bm{\bar g}_{\bm{\theta}_{q(k)}}^{t}\Big\lVert_{2}^{2}\\
	\le & L^{2}\sum\limits_{k=1}^{K}\beta^{K-k}\Big\lVert\bm{\theta}_{q(k)}^{t} - \bm{\theta}_{q(k)}^{d_{k,t}}\Big\lVert_{2}^{2}\\
	=& L^{2}\sum\limits_{k=1}^{K}\beta^{K-k}\Big\lVert\sum\limits_{j=\mathrm{max}\{0,d_{k,t}\}}^{t-1}(\bm{\theta}_{q(k)}^{j+1} - \bm{\theta}_{q(k)}^{j})\Big\lVert_{2}^{2}\\
	\le & L^{2}\sum\limits_{k=1}^{K}\beta^{K-k}\sum\limits_{j=\mathrm{max}\{0,d_{k,t}\}}^{t-1}\Big\lVert\bm{\theta}_{q(k)}^{j+1} - \bm{\theta}_{q(k)}^{j}\Big\lVert_{2}^{2}\\
	= & L^{2}\sum\limits_{k=1}^{K}\beta^{K-k}\sum\limits_{j=\mathrm{max}\{0,d_{k,t}\}}^{t-1}\gamma_{j}^{2}\beta^{2(K-k)}\Big\lVert\bm{g}_{\bm{\theta}_{q(k)}}^{d_{k,t}}\Big\lVert_{2}^{2}\\
	\le & L^{2}M\sum\limits_{k=1}^{K}\beta^{3(K-k)}\sum\limits_{j=\mathrm{max}\{0,d_{k,t}\}}^{t-1}\gamma_{j}^{2}\\
	\le &  \gamma_{t}^{2}L^{2}M\sum\limits_{k=1}^{K}\beta^{3(K-k)}(t-\mathrm{max}\{0,d_{k,t}\})
	\end{align*}
	with the first inequality coming from the Assumption 1. On the other hand, The expectation of $\tilde{Q}_2$ is bounded by
	
	\begin{align*}
	\mathrm{E}[\tilde{Q}_2 ]=&  -(\gamma_{t} - L\gamma_{t}^{2})\mathrm{E}\Big[\sum\limits_{k=1}^{K}\beta^{K-k}(\bm{\bar g}_{\bm{\theta}_{q(k)}}^{t})^{T}\Big(\bm{g}_{\bm{\theta}_{q(k)}}^{d_{k,t}} - \bm{\bar g}_{\bm{\theta}_{q(k)}}^{t}\Big)\Big]\\
	=&-(\gamma_{t} - L\gamma_{t}^{2})\sum\limits_{k=1}^{K}\beta^{K-k}(\bm{\bar g}_{\bm{\theta}_{q(k)}}^{t})^{T}\Big(\bm{\bar g}_{\bm{\theta}_{q(k)}}^{d_{k,t}} - \bm{\bar g}_{\bm{\theta}_{q(k)}}^{t}\Big)\\
	\le & \frac{\gamma_{t} - L\gamma_{t}^{2}}{2}\sum\limits_{k=1}^{K}\beta^{K-k}\Big\lVert\bm{\bar g}_{\bm{\theta}_{q(k)}}^{t}\Big\lVert_{2}^{2} + \frac{\gamma_{t} - L\gamma_{t}^{2}}{2}\tilde{P}_1
	\end{align*}
	where the second equality follows by the unbiased gradient using SGD, and the inequality comes from $\pm \bm{x}^{T}\bm{y}\le \frac{1}{2}\lVert\bm{x}\lVert_2^2 + \frac{1}{2}\lVert\bm{y}\lVert_2^2$.
	
	Taking the expectation of both sides in \eqref{eq_appendix_3} and substituting $\tilde{Q}_1$ and $\tilde{Q}_2$, the inequality is rewritten as
	\begin{align*}
	\mathrm{E}\Big[f(\bm{\theta}^{t+1})\Big] \le& \mathrm{E}\Big[f(\bm{\theta}^{t})\Big]- (\gamma_{t} - \frac{L\gamma_{t}^{2}}{2})\sum\limits_{k=1}^{K}\beta^{K-k}\Big\lVert\bm{\bar g}_{\bm{\theta}_{q(k)}}^{t}\Big\lVert_2^2+\frac{\gamma_{t} - L\gamma_{t}^{2}}{2}\sum\limits_{k=1}^{K}\beta^{K-k}\Big\lVert\bm{\bar g}_{\bm{\theta}_{q(k)}}^{t}\Big\lVert_{2}^{2} + \frac{\gamma_{t} - L\gamma_{t}^{2}}{2}\tilde{P}_1 \\
	&+ L\gamma_{t}^{2}M\frac{\beta^{K}-1}{\beta - 1} + L\gamma_{t}^{2}\tilde{P}_1\\
	\le&\mathrm{E}\Big[f(\bm{\theta}^{t})\Big]- \frac{\gamma_{t}}{2}\sum\limits_{k=1}^{K}\beta^{K-k}\Big\lVert\bm{\bar g}_{\bm{\theta}_{q(k)}}^{t}\Big\lVert_2^2 + \frac{\gamma_{t}+ L\gamma_{t}^{2}}{2}\gamma_{t}^{2}L^{2}M\sum\limits_{k=1}^{K}\beta^{3(K-k)}(t-\mathrm{max}\{0,d_{k,t}\})+ L\gamma_{t}^{2}M\frac{\beta^{K}-1}{\beta - 1} \\
	=&\mathrm{E}\Big[f(\bm{\theta}^{t})\Big] - \frac{\gamma_{t}}{2}\sum\limits_{k=1}^{K}\beta^{K-k}\Big\lVert\bm{\bar g}_{\bm{\theta}_{q(k)}}^{t}\Big\lVert_2^2+ \gamma_{t}^{2}\Big( LM\frac{\beta^{K}-1}{\beta - 1} + \frac{\gamma_{t}+ L\gamma_{t}^{2}}{2}L^{2}M\sum\limits_{k=1}^{K}\beta^{3(K-k)}(t-\mathrm{max}\{0,d_{k,t}\})\Big)\\
	\le&\mathrm{E}\Big[f(\bm{\theta}^{t})\Big] - \frac{\gamma_{t}}{2}\sum\limits_{k=1}^{K}\beta^{K-k}\Big\lVert\bm{\bar g}_{\bm{\theta}_{q(k)}}^{t}\Big\lVert_2^2+ \gamma_{t}^{2}\Big( LM\frac{\beta^{K}-1}{\beta - 1} + LM\sum\limits_{k=1}^{K}\beta^{3(K-k)}(t-\mathrm{max}\{0,d_{k,t}\})\Big)
	\end{align*}
	where the last inequality follows from $L\gamma_{t}\le 1$ such that $\frac{\gamma_{t}+ L\gamma_{t}^{2}}{2}L\le 1$. The proof is now completed
\end{proof}	

\end{document}